\documentclass{article}
\pdfoutput=1
\usepackage{ijcai21}

\usepackage{times}

\usepackage{soul}
\usepackage{url}
\usepackage[hidelinks]{hyperref}
\usepackage[utf8]{inputenc}
\usepackage[small]{caption}
\usepackage{graphicx}
\usepackage{amsmath}
\usepackage{booktabs}
\usepackage{amssymb} 
\usepackage{color}
\usepackage{subfigure}

\usepackage{relsize}
\usepackage{mathtools}
\usepackage{enumitem}
\usepackage{multirow,multicol}

\newtheorem{definition}{Definition}[section]

\newcommand\ChangeRT[1]{\noalign{\hrule height #1}}

\usepackage{etoolbox,siunitx}
\usepackage{tabularx}
\usepackage{array}
\usepackage{pgf, tikz}
\usepackage{pgfplots}

\usepackage{tcolorbox}
\let\oldbibliography\thebibliography
\renewcommand{\thebibliography}[1]{\oldbibliography{#1}
\setlength{\itemsep}{0pt}} 

\DeclareMathSizes{7}{7}{5}{4}

\DeclareMathOperator{\A}{\mathbf{A}}
\DeclareMathOperator{\An}{\Tilde{\mathbf{A}}}

\DeclareMathOperator{\Ll}{\mathbf{L}}
\DeclareMathOperator{\Ln}{\Tilde{\mathbf{L}}}

\DeclareMathOperator{\D}{\mathbf{D}}
\DeclareMathOperator{\Dn}{\Tilde{\mathbf{D}}}
\DeclareMathOperator{\Drw}{\mathbf{D}^{-1}}
\DeclareMathOperator{\Dsym}{\mathbf{D}^{-\frac{1}{2}}}

\DeclareMathOperator{\Dnsym}{\Tilde{\mathbf{D}}^{-\frac{1}{2}}}

\DeclareMathOperator{\Z}{\mathbf{Z}}
\DeclareMathOperator{\X}{\mathbf{X}}
\DeclareMathOperator{\h}{\mathbf{h}}

\DeclareMathOperator{\N}{\mathcal{N}}

\DeclareMathOperator{\Pp}{\mathbf{P}}
\DeclareMathOperator{\Qq}{\mathbf{Q}}
\DeclareMathOperator{\I}{\mathbf{I}}

\DeclareMathOperator{\Uu}{\mathbf{U}}
\DeclareMathOperator{\UT}{\mathbf{U^{\intercal}}}
\DeclareMathOperator{\uu}{\mathbf{u}}
\DeclareMathOperator{\ut}{\mathbf{u^{\intercal}}}
\DeclareMathOperator{\g}{\mathbf{g}}
\DeclareMathOperator{\LBD}{\mathbf{\Lambda}}

\title{Bridging the Gap between Spatial and Spectral Domains: \\A Survey on Graph Neural Networks}

\author{
Zhiqian Chen$^1$\and
Fanglan Chen$^2$\and
Lei Zhang$^2$\and
Taoran Ji$^2$\and
Kaiqun Fu$^2$\and\\
Liang Zhao$^4$\and
Feng Chen$^3$\and
Lingfei Wu$^6$\and
Charu Aggarwal$^5$\And
Chang-Tien Lu$^2$\\
\affiliations
$^1$Mississippi State University,
$^2$Virginia Tech\\
$^3$University of Texas at Dallas,
$^4$Emory University,
$^5$IBM Research,
$^6$JD.COM
\emails
zchen@cse.msstate.edu,
\{fanglanc, zhanglei, jtr, fukaiqun,ctlu\}@vt.edu,
feng.chen@utdallas.edu,
liang.zhao@emory.edu,
lingfei.wu@jd.com,
charu@us.ibm.com
}

\begin{document}

\maketitle

\begin{abstract}
Deep learning's success has been widely recognized in a variety of machine learning tasks, including image classification, audio recognition, and natural language processing. 
As an extension of deep learning beyond these domains, graph neural networks (GNNs) are designed to handle the non-Euclidean graph-structure which is intractable to previous deep learning techniques. 
Existing GNNs are presented using various techniques, making direct comparison and cross-reference more complex. 
Although existing studies categorize GNNs into spatial-based and spectral-based techniques, there hasn't been a thorough examination of their relationship. 
To close this gap, this study presents a single framework that systematically incorporates most GNNs. 
We organize existing GNNs into spatial and spectral domains, as well as expose the connections within each domain. A review of spectral graph theory and approximation theory builds a strong relationship across the spatial and spectral domains in further investigation.
\end{abstract}

\section{Introduction}

The effectiveness of deep learning \cite{lecun2015deep} has been widely recognized in various machine learning tasks \cite{redmon2016you,hinton2012deep,luong2015effective}, especially on the Euclidean data. Recent decades have witnessed a great number of emerging applications where effective information analysis generally boils down to the non-Euclidean geometry of the data represented by a graph, including social networks, transportation networks, disease contact networks, brain's neuronal networks, gene data on biological regulatory networks, telecommunication networks, and knowledge graph~\cite{zhang2018deep,zhou2018graph,wu2019comprehensive,hamilton2017representation}. General deep learning techniques can hardly handle such non-Euclidean problems in graph-structured data.
Representing a graph is challenging because it is irregular, i.e., each graph has a variable size of nodes, and each node in a graph has a different number of neighbors, rendering some operations such as convolutions not compatible with the graph structure. 
Deep learning-based approaches for graph data have recently piqued people's curiosity.
This growing trend has attracted the attention of the machine learning community, and a huge number of GNN models based on various theories have been constructed. \cite{bruna2014spectral,kipf2016semi,defferrard2016convolutional,hamilton2017inductive,atwood2016diffusion,velivckovic2017graph}.

However, there is still a lack of comprehension of GNN's physical meaning and representational capacity. This constraint not only makes it difficult to compare and enhance state-of-the-art procedures, but it also makes it an unmanageable black-box that can pose serious concerns in areas such as business intelligence and drug research.
As a result, there is a compelling need to de-mystify GNNs, prompting researchers to explain GNNs in a broader context \cite{xu2018how,gilmer2017neural,ying2019gnn,yuan2020explainability}. However, these studies can only explain a small number of GNNs, leaving the majority of them unsolved.

Spatial-based methods (e.g., random walk, Page Rank, attention model, low-pass filter, message passing) and spectral-based methods (e.g., random walk, Page Rank, attention model, low-pass filter, message passing) are the two types of GNNs currently in use (e.g., ARMA filter, auto-regressive filter). 
Existing surveys \cite{bronstein2017geometric,zhang2018deep,zhou2018graph,wu2019comprehensive} have done a good job of separating and summarizing the works in their various subcategories. 
However, under current taxonomies, comprehension of the linkages between spatial and spectral-based techniques has not been fully explored. 
GNN research is still focused on revealing the underlying mechanisms, which is difficult due to the fact that most GNN mechanisms are not intrinsically consistent \cite{yuan2020explainability,xu2018how,Li_2019_CVPR,li2018deeper}.

The goal of this research is to propose a generic framework for unifying GNNs under various processes by uncovering hidden relationships between GNNs from a theoretical standpoint. Our study is unusual in that it connects current GNN research in a white-box, interpretable framework. 
To begin, we provide a quick overview of the suggested framework for connecting the spatial and spectral domains, as well as their linkages. Then, for each subdivision of the spatial and spectral domains, many sample works are presented. Finally, we show how current work on enhancing GNNs in terms of over-smoothing and how a large scale graph can be broken down into one of these subcategories.
Detailed contributions of this paper are summarized as follows: 
    
        
\begin{enumerate}[leftmargin=*]
    \item \textbf{Proposing taxonomies for the spatial- and spectral-based methods, respectively.}
    This paper unifies GNNs in the spatial domain by formulating node aggregation,
    and categorizes GNNs by the types of frequency response functions in the spectral domain.

    \item \textbf{Unifying spatial and spectral models into a generic framework.} 
    The proposed framework links the spectral and spatial domain by 
    comparing the analytical forms of \textit{node aggregation function} and \textit{frequency response function}. 
    
    \item \textbf{Analyzing the proposed framework with trendy topics.}
    Time complexity and expressive power are studied with approximation theory. Also,
    over-smoothing and scaling issue, two severe problems in frontier research, are studied, showing that they are special cases of our proposed framework.
    
\end{enumerate}

\begin{figure*}[!t]
    \centering
    \includegraphics[width=1.0\linewidth]{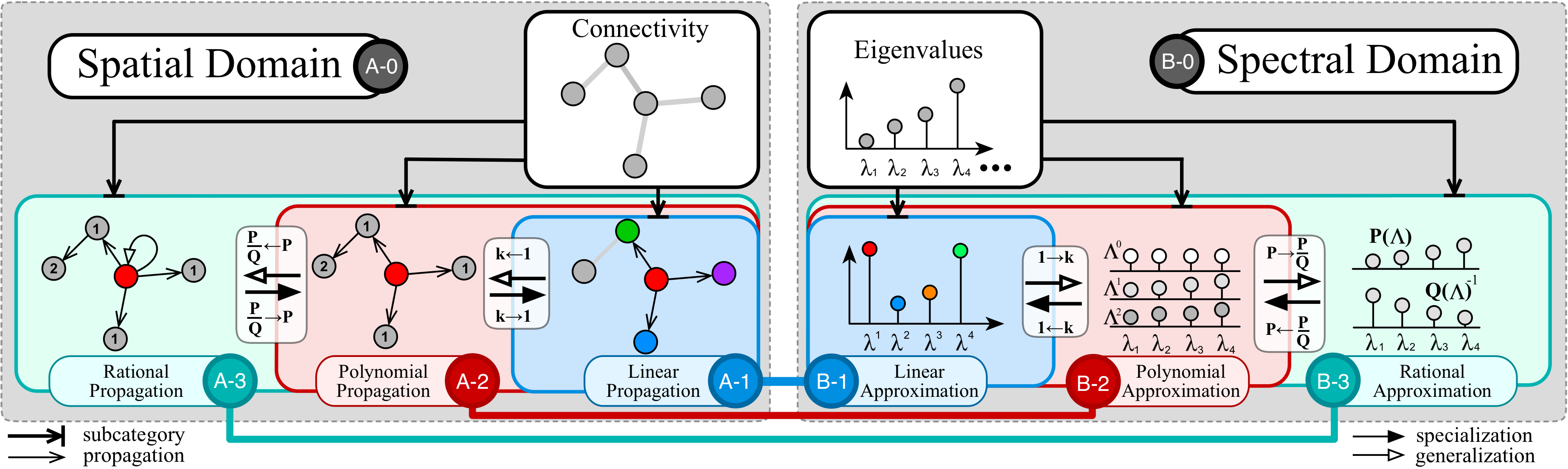}
    \caption{Illustration of major graph neural operations and their relationship. Spatial and spectral methods are divided into three groups, respectively. Group A-1, A-2, and A-3 are strongly related by generalization and specialization, so are group B-1, B-2, and B-3. The equivalence relationships between A-1 and B-1 are marked with blue, red for A-2/B-2, and green for A-3/B-3.}
    \label{fig:overview}
\end{figure*} 
\section{Problem Setup and Framework Overview}\label{sec:problem}

A graph is defined as $\mathcal{G} = (\mathcal{V}, \mathcal{E}, \A)$, where $\mathcal{V}$ is a set of n nodes and $\mathcal{E}$ represents edges.
An entry $v_{i} \in \mathcal{V}$ denotes a node, $e_{i,j}={\{v_{i}, v_{j}\}} \in \mathcal{E}$ indicates an edge between nodes $i$ and $j$.
The adjacency matrix $\A\in \mathbb{R}^{N\times N}$ is defined by if $\A_{i,j} =1$ 
there is a link between node $i$ and $j$, and else 0. Node features $\X \in \mathbb{R}^{N\times F}$ is a matrix with each entry $x_{i}\in\X$ representing the feature vector on node $i$. Another popular graph matrix is the graph Laplacian which is defined as $\Ll= \D-\A \in \mathbb{R}^{N\times N}$ where $\D$ is the degree matrix. Due to its generalization ability \cite{bollobas2004extremal}
, the symmetric normalized Laplacian is often used, which is defined as $\Ln=\D^{-\frac{1}{2}}\Ll\D^{-\frac{1}{2}}$. Another option is random walk normalization: $\Ln=\D^{-1}\Ll$. 
Note that normalization could also be applied to the adjacency matrix. Their relationship can be derived as $\Ln= \I-\An$.
Since this survey focuses on spectral- and spatial-based methods, two related definitions are listed below.
\begin{definition}[Spatial Method]
By integrating graph connectivity $\mathcal{G}$ and node features $\X$, the updated node representations ($\Z$) are defined as:
\begin{equation}
\Z=f(\mathcal{G})\X,
\end{equation}where $\mathcal{G}$ is often implemented with $\A$ or $\Ll$ in existing works. Therefore, spatial methods focus on finding a \textbf{node aggregation function} $f(\cdot)$ that learns how to aggregate node features to obtain a updated node embedding $\Z$.
\label{def:spatial}
\end{definition}

The Laplacian $\Ll$ can be diagonalized by the Fourier basis $\UT$ (i.e., graph Fourier transform) \cite{shuman2013emerging,zhu2012approximating}: $\Ln = \Uu \LBD \UT$, where $\LBD$ is the diagonal matrix whose diagonal elements are the corresponding eigenvalues (i.e., ${\displaystyle \LBD_{ii}=\lambda _{i}}$), and $\Uu$ is also called eigenvectors.
The graph Fourier transform of a signal $\X$ is defined as $\hat{\X}=\UT \X \in \mathbb{R}^{N\times N}$ and its inverse as $\X=\Uu \hat{\X}$. 
\begin{definition}[Spectral Method]
A graph convolution operation is defined in the Fourier domain such that 
\begin{equation*}\small
f_{1}*f_{2}=\Uu \left[\left(\UT f_{1} \right) \odot \left(\UT f_{2}\right)\right],
\end{equation*}where $\odot$ is the element-wise product, and $f_{1}/f_{2}$ are two signals defined on nodes. It follows that a node
signal $f_{2}=\X$ is filtered by spectral signal $\hat{f_{1}}=\UT f_{1}=\g$ as:
\begin{equation}\small
\Z=\g(\Ln)\X= \Uu \left[\g(\LBD)\odot \left(\UT\X\right)\right] = \Uu \g(\LBD) \UT \X,
\end{equation} where $\g$ is known as \textbf{frequency response function}. Therefore, the objective of spectral methods is to learn a function $\g(\cdot)$.
\label{def:spectral}
\end{definition}

\vspace{2pt}
\noindent\textbf{Framework Overview}:
As shown in Fig. \ref{fig:overview}, the proposed framework categorizes GNNs into the spatial (A-0) and spectral (B-0) domains, either of which is further divided into three subcategories (A-1/A-2/A-3 and B-1/B-2/B-3), respectively. Based on the types of node aggregation (i.e., $f$ in Def. \ref{def:spatial}), A-0 is further divided into linear (A-1), polynomial (A-2) and rational (A-3) propagation. Linear propagation means only the first order neighbors are involved, while polynomial propagation considers high-order neighbors. Beyond them, rational propagation adds reverse propagation.
B-0 is split into linear (B-1), polynomial (B-2) and rational (B-3) approximation based on the types of frequency filtering (i.e., $\g$ in Def. \ref{def:spectral}), since they explicitly belong to the approximation techniques.
Each category and subcategory will be elaborated with examples in Section 3 and 4.
Two major relationships are inside our framework:
\paragraph{(1) \underline{Inter}-Relationship between A-0 and B-0}\mbox{}\\
By definition \ref{def:spatial} and \ref{def:spectral}, there is a correspondence between node aggregation function ($f$) and frequency response function ($\g$):
\begin{equation*}
    f(\mathcal{G})\X=\Uu \g(\LBD) \UT \X,
\end{equation*}
where $f=\g$ and three pairs of equivalence relationships exist across A-0 and B-0:
\begin{itemize}
    \item \textbf{{Linear Propagation} and {Approximation}}: Through graph and matrix theories, A-1 adjusts weights on a set of neighbor nodes, which corresponds to adjusting weights on frequency components in B-1.
    \item \textbf{{Polynomial Propagation} and {Approximation}:} Aggregating different orders of neighbors in A-2 can be rewritten as the sum of different orders of frequency components, which is exactly the analytical form of B-2.
    \item \textbf{{Rational Propagation} and {Approximation}:} A-3 defines a label propagation with reverse propagation, which is equivalent to B-3 that approximates frequency response with rational function.
\end{itemize}

\paragraph{(2) \underline{Intra}-Relationship inside A-0 and B-0}\mbox{}\\
Three spatial subcategories (A-1/A-2/A-3) are strongly connected under \textbf{(1)}\textit{ Generalization}:
A-1 can be extended to A-2 by adding more neighbors of higher order. A-2 can be upgraded to A-3 by adding reverse propagation; 
\textbf{(2)}\textit{ Specialization:}
A-1 is a special case of A-2 when setting the order to 1. A-2 is a special case of A-3 if the reverse propagation is removed.

Similarly, three spectral subcategories (B-1/B-2/B-3) are strongly connected with
\textbf{(1)}\textit{ Generalization}: B-1 can be extended to B-2 by adding more higher order of eigenvalues. B-2 can be upgraded to B-3 if the denominator of frequency response function is not 1;
\textbf{(2)}\textit{ Specialization}:
B-1 is a special case of B-2 by setting the highest order to 1. B-2 is a special case of B-3 by setting the denominator of the frequency response function to 1.

\section{Spatial-based GNN (A-0)} \label{sec:spatial}
According to the node aggregation, we categorize spatial-based GNNs into three subcategories below, which have different strategies on the neighbor selection and propagation flow.
    
\subsection{Linear Propagation (A-1)}\label{sec:spatial_a1}
Many works \cite{perozzi2014deepwalk,xu2018how,gilmer2017neural,hamilton2017inductive,velivckovic2017graph} can be treated as learning the aggregation scheme among first-order neighbors (i.e., direct neighbors). This aspect focuses on learning the weights for the node and its direct neighbors. Formally, updated node embeddings, $\Z(v)$, can be written as:
\begin{equation*}\small
    \Z(v_{i}) = \Phi(v_{i}) \h(v_{i}) + \sum_{u_{j}\in\mathcal{N}(v_{i})}\Psi(u_{j})\h(u_{j}), 
\label{eq:a1}
\end{equation*}where $u_{j}$ denotes a neighbor of node $v_{i}$, $\h(\cdot)$ is their representations, and $\Phi/\Psi$ indicate the weight functions. First item on the right hand side denotes the representation of node $v_{i}$, while the second term represents the update from its neighbors. Applying random walk normalization, Eqn. \ref{eq:a1} can be written as:
\begin{equation*}\small
    \Z(v_{i}) = \Phi(v_{i}) \h(v_{i}) +  
    \sum_{u_{j}\in\mathcal{N}(v_{i})}\Psi(u_{j})\frac{\h(u_{j})}{d_{i}},
\label{eq:a1_comb}
\end{equation*}or with symmetric normalization:
\begin{equation*}\small
    \Z(v_{i}) = \Phi(v_{i}) \h(v_{i}) +  
    \sum_{u_{j}\in\mathcal{N}(v_{i})}\Psi(u_{j})\frac{\h(u_{j})}{\sqrt{d_{i}d_{j}}},
\label{eq:a1_sym}
\end{equation*}
where $d_{i}$ represents the degree of node $v_{i}$. 
Normalization has better generalization capacity, which is supported by a theoretical proof on performance improvement \cite{johnson2007effectiveness}. In a simplified configuration, weights for the neighbors ($\Psi$) are the same. Therefore, they can be rewritten in matrix form as:
\begin{equation}\small
    \Z = \phi \X + \psi\D^{-1}\A\X  = (\phi \I + \psi\D^{-1}\A)\X, 
\label{eq:a1_mat_comb}
\end{equation} or 
\begin{equation}\small
\Z=\phi\X + \psi\Dsym\A\Dsym\X = (\phi \I + \psi\Dsym\A\Dsym)\X,
\end{equation}where $\phi$ and $\psi$ are the weights. All the above can be generalized as the same form:
\begin{equation}\small
        \Z =(\phi \I + \psi\tilde{\A})\X,
        \label{eq:a1_final}
\end{equation}
where $\tilde{\A}$ denotes the normalized $\A$, which could be implemented by random walk or symmetric normalization. 
Several state-of-the-art models are selected as examples in this category: 

\vspace{5pt}
\noindent\textbf{(1) Graph Convolutional Network (GCN)}~\cite{kipf2016semi} adds a self-loop to the current node , and applies a \textit{renormalization}, changing $\D_{ii}=\sum_{j}\A_{ij}$ to $\hat{\D}_{ii}=\sum_{j}(\A+\I)_{ij}$. GCN can be formulated as:
\begin{equation*}\small
\Z = \hat{\D}^{-\frac{1}{2}}\hat{\A}\hat{\D}^{-\frac{1}{2}}\X = \hat{\D}^{-\frac{1}{2}}(\I+\A)\hat{\D}^{-\frac{1}{2}} \X=(\I+\An)\X,
\label{eq:a1_gcn}
\end{equation*}
where $\hat{\A}=\A+\I$.
Therefore, GCN is equivalent to Eqn. \ref{eq:a1_final} when setting $\phi=0$ and $\psi=1$ with the renormalization trick, and the result of GCN is exactly the sum of the current node and average of its neighbors. 

\vspace{5pt}
\noindent\textbf{(2) GraphSAGE}~\cite{hamilton2017inductive} with mean aggregator averages a node with its neighbors by:
\begin{equation}\small
\Z(v_{i}) =\operatorname{MEAN}\left(\left\{\h(v_{i})\right\} \cup\left\{\h(u_{j}), \forall u_{j} \in \mathcal{N}(v_{i})\right\}\right), 
\label{eq:a1_sage_intro}
\end{equation}where $\h$ indicates the representation, and $\mathcal{N}$ denotes the neighbor nodes. Eqn. \ref{eq:a1_sage_intro} can be written in matrix form as:
\begin{equation}\small
\Z = \hat{\D}^{-1}(\I+\A)\X=(\hat{\D}^{-1}+\An)\X,
\label{eq:a1_sage}
\end{equation} 
which is equivalent to Eqn. \ref{eq:a1_final} with $\phi=1$ and $\psi=1$. Note that the key difference between GCN and GraphSAGE is the normalization type. The former is symmetric normalization and the latter is random walk normalization.

\vspace{5pt}
\noindent\textbf{(3) Graph Isomorphism Network (GIN)}~\cite{xu2018how} updates node representations as:
\begin{equation}
\small
    \Z=\left(1+\epsilon\right) \cdot \h(v)+\sum_{u_{j} \in \mathcal{N}(v_{i})} \h_(u_{j})=[(1+\epsilon)\I + \A]\X,
\label{eq:a1_gin}
\end{equation} 
which is equivalent to Eqn. \ref{eq:a1_final} with $\phi=1+\epsilon$ and $\psi=1$. Note that \textbf{GIN} dose not perform normalization.

\subsection{Polynomial Propagation (A-2)}\label{sec:spatial_a2}
To collect richer local structure, numerous studies involve high-order neighbor nodes (i.e., neighbors of neigbhors)~\cite{atwood2016diffusion,defferrard2016convolutional,wu2019simplifying,tang2015line,grover2016node2vec}, since direct neighbors (i.e., first-order neighbors) are not always sufficient for representing the node. On the other hand, if the order number is large, GNNs may average all node representations, causing an over-smoothing issue and losing its focus on the local neighborhood~\cite{li2018deeper}. This motivates many models to tune the aggregation scheme on different orders of neighbors. Therefore, proper constraint and flexibility of orders are critical for node representations. High-order neighbors have been proved to characterize complex signal such as Gabor-like filters \cite{abu2019mixhop}.
Formally, this type of work can be formulated as:
{\scriptsize
\begin{align}
    \Z(v_{i})& = \phi \h(v) 
    +\overbrace{\sum_{u_{j}^{(1)}\in \N(v)} \psi_{j}^{(1)} \h(u_{j}^{(1)})}^\text{1st order neighbors} +
    \overbrace{\sum_{u_{j}^{(2)}\in \bigcup_{j} \N(u_{j}^{(1)})} \psi_{j}^{(2)} \h(u_{j}^{(2)})}^\text{2nd order neighbors} &\nonumber
    \\
    &...+
    \overbrace{\sum_{u_{j}^{(k+1)}\in \bigcup_{j} \N(u_{j}^{(k)})} \psi_{j}^{(k+1)} \h(u_{j}^{(k+1)})}^\text{k-th order neighbors}+...,
\label{eq:a2}
\end{align}
}where $u^{(k)}_{i}$ indicates a $k$-th order neighbors of node $v$ and $\N(v)$ denotes direct neighbors of $v$. Eqn. \ref{eq:a2} can be rewritten in matrix form as:
\begin{equation}\small
        \Z= (\phi\I + \sum_{j=1}^{k} \psi_{j}\A^{j}) \X = \Pp(\A)\X,
\label{eq:a2_mat}
\end{equation}
where $\Pp(\cdot)$ is a polynomial function. Applying normalization, Eqn. \ref{eq:a2_mat} can be rewritten as:
{\small
\begin{align}
\Z &= (\phi \I + \sum_{j=1}^{k}\psi_{i} (\Dsym\A\Dsym)^{j})\X 
    =\Pp(\An)\X,&
\label{eq:a2_mat_norm}
\end{align}
}where $\phi = \psi_{0}$, and $\A$ could also be normalized by random walk normalization. Several existing works are analyzed below, showing that they are variants of Eqn. \ref{eq:a2_mat} or \ref{eq:a2_mat_norm}:

\vspace{5pt}
\noindent\textbf{(1) ChebNet}~\cite{hammond2011wavelets} 
introduced truncated Chebyshev polynomial for estimating wavelet in graph signal processing. Based on this polynomial approximation, Defferrard et al.~\cite{defferrard2016convolutional} designed ChebNet which embeds a novel neural network layer for the convolution operator. Specifically, ChebNet is written as:
\begin{equation}\small
    \sum_{k=0}^{K-1} \theta_{k} T_{k}(\tilde{\Ll}) \X= (\tilde{\theta_{0}}\I + \tilde{\theta_{1}}\tilde{\Ll} + \tilde{\theta_{2}}\tilde{\Ll}^{2} +... ) \X,
\end{equation}
where $T_{k}(\cdot)$ denotes the Chebyshev polynomial and $\theta_{k}$ is the Chebyshev coefficient. $\tilde{\theta}$ is the coefficient after expansion and reorganization. Since $\Ln=\I-\An$, we have:
\begin{equation}\small
   \sum_{k=0}^{K-1} \theta_{k} T_{k}(\tilde{\Ll}) \X=  [\tilde{\theta_{0}}\I + \tilde{\theta_{1}}(\I-\An) + \tilde{\theta_{2}}(\I-\An)^{2} +... ] \X,
\end{equation}
which can be reorganized as:
\begin{equation}\small
  \sum_{k=0}^{K-1} \theta_{k} T_{k}(\tilde{\Ll}) \X= (\phi \I +\sum_{i=1}^{k} \psi_{i} \An^{i})\X =\Pp(\An)\X,
\end{equation} which is exactly Eqn. \ref{eq:a2_mat_norm}.

\vspace{5pt}
\noindent\textbf{(2) Diffusion Convolutional Neural Network (DCNN)}~\cite{atwood2016diffusion} consider using a degree-normalized transition matrix, i.e., renormalized adjacency matrix $\An = \Dn\A$:
\begin{equation*}\small
    \Z=W \odot \An^{*} \X,
\end{equation*}where $\An^{*}$ denotes a tensor containing the power series of $\An$, and the $\odot$ operator represents element-wise multiplication. It can be transformed as:
\begin{equation}\small
  \Z=(\psi_{1}\An + \psi_{2}\An^{2}+ \psi_{3}\An^{3} +...)\X=\Pp(\An)\X,
\end{equation}which equals to Eqn. \ref{eq:a2_mat_norm}.

\vspace{5pt}
\noindent\textbf{(3) Simple Graph Convolution (SGC)}~\cite{wu2019simplifying} removes non-linear function between neighboring graph convolution layers, and combines graph propagation in one single layer:
\begin{equation}\small
    \Z = \An^{K}\X,
\end{equation} where $\An$ is renormalized adjacency matrix, i.e., $\An= \Dnsym\A\Dnsym$, and $\Dnsym$ is degree matrix with self loop (same as in GCN). Therefore, it can be rewritten as:
\begin{equation}\small
    \Z = (0\cdot\I + 0\cdot \An + 0\cdot\An^{2} +...+1\cdot\An^{K})\X = \Pp(\An) \X.
\end{equation}



\subsection{Rational Propagation (A-3)}\label{sec:spatial_a3}
Most works merely consider label propagation to neighbors, ignoring propagation in the reverse direction. Reverse propagation on labels or features alleviates the over-smoothing issue by propagating back to the current node or restarting propagation with a certain probability. Note that A-2 can also mitigate over-smoothing issue by manually adjusting the order number, while A-3 can do it automatically.
Several works explicitly or implicitly implement reverse propagation by applying rational function on the adjacency matrix \cite{klicpera2018predict,wu2012learning,Li_2019_CVPR,7131465,7581108,levie2018cayleynets,bianchi2019graph,chen2018rational}. Formally, this subcategory can be represented as:
\begin{equation}\small
    \Z =  \Pp(\An)\Qq(\An)^{-1}\X  =\frac{\Pp(\An)}{\Qq(\An)}\X,
\label{eq:a3_mat}
\end{equation}where $\Pp$ and $\Qq$ are two different polynomial functions, and the bias of $\Qq$ is often set to 1.

\vspace{5pt}
\noindent\textbf{(1) Auto-Regressive Label Propagation}~\cite{zhu2003semi,zhou2004learning,bengio200611} is a widely used methodology for graph-based learning. The objective of Label Propagation (LP) is two-fold: one is to extract embeddings that match with the label, the other is to become similar to neighboring nodes. The label can be treated as part of node features, so we have:
\begin{equation*}\small
    \Z =( \I + \alpha \Ln )^{-1} \X = \frac{\I }{\I +\alpha (\I -\An )} \X = \frac{\I }{(1+ \alpha )\I - \alpha \An} \X,
\end{equation*}
which is equivalent to the form of Eqn. \ref{eq:a3_mat}, i.e., $\Pp=\I$ and $\Qq=(1+\alpha)\I- \alpha \An$.

\vspace{5pt}
\noindent\textbf{(2) Personalized PageRank (PPNP)}~\cite{klicpera2018predict} obtains node's representation via a teleport or restart probability $\alpha$ which is the ratio of keeping the original representation $\X$, i.e., reverse or no propagation. (1-$\alpha$) is the ratio of performing the normal label propagation:
\begin{equation*}\small
    \Z=\alpha\left(\I-(1-\alpha) \An\right)^{-1} \X = \frac{\alpha}{\I-(1-\alpha) \An} \X,
\label{eq:ppnp}
\end{equation*} where $\An=\Drw\A$ is random walk normalized adjacency matrix with self loop. 

\vspace{5pt}
\noindent\textbf{(3) ARMA filter}~\cite{bianchi2019graph} utilizes ARMA filter for approximating the desired filter response function, which can be written in the spatial domain as:
\begin{equation*}\small
    \Z=\frac{b}{1-a\An}\X.
\end{equation*} Note that ARMA filter is an unnormalized version of PPNP. When setting $a+b=1$, ARMA equals to PPNP.

\section{Spectral-based GNN (B-0)} \label{sec:spectral}
Spectral-based GNN models are built on spectral graph theory which applies eigen-decomposition and analyzes the weight-adjusting function (i.e., frequency response function) on eigenvalues of graph matrices. Based on spectral operation, we propose a new taxonomy of GNNs, categorizing spectral-based GNN into three subcategories below, and use the same GNN examples in Section \ref{sec:spatial}. 

\begin{table}[]
\centering
\scalebox{0.92}{
\begin{tabular}{|l|c|c|}\ChangeRT{2pt}
                                          & GNN Model   & frequency response function \\ \ChangeRT{1pt}
\multicolumn{1}{|l|}{\multirow{3}{*}{B1-Linear}} & GCN     & $1-\LBD$ \\ \cline{2-3} 
\multicolumn{1}{|l|}{}                    & GraphSAGE    & $2-\LBD$ \\ \cline{2-3} 
\multicolumn{1}{|l|}{}                    & GIN     & $1+\epsilon+\LBD$ \\ \ChangeRT{1pt}
\multirow{3}{*}{B2-Polynomial}                       & ChebNet & $\tilde{\theta_{0}}\cdot 1 + \tilde{\theta_{1}}\LBD + \tilde{\theta_{2}}{\LBD}^{2} +...$  \\ \cline{2-3} 
                                          & DCNN    & $\psi_{1}\cdot 1 + \psi_{2}\cdot \LBD+ \psi_{3}\LBD^{3} +...$  \\ \cline{2-3} 
                                          & SGC     & $ \sum_{i}^{n}\left(\begin{array}{c}{K} \\ {i}\end{array}\right)  \LBD^{i}$\\ \ChangeRT{1pt}
\multirow{3}{*}{B3-Rational}                       & AR      &  $\frac{1}{1+\alpha (1-\LBD)}$\\ \cline{2-3} 
                                          & PPNP    &  $\frac{\alpha}{\alpha+(1-\alpha) \LBD}$ \\ \cline{2-3} 
                                          & ARMA    &  $\frac{b}{1-a+a\LBD}$                 \\ \ChangeRT{2pt}
\end{tabular}}
\caption{Frequency response functions grouped by approximation theory. Parameter notations have been defined in section \ref{sec:spatial}.}
\label{tab:freq_func}
\end{table}

\subsection{Linear Approximation (B-1)}\label{sec:spectral_b1}
Many existing works are proved to be low-pass filters \cite{Li_2019_CVPR}, which means that only low-frequency components are emphasized.
All the works that can be classified into A-1 can be understood as adjusting weights of frequency component during aggregation. Specifically, a linear function of $\g$ is defined so that:
\begin{equation}\small
    \Z = (\sum_{i=0}^{l} \theta_{i}\lambda_{i} \uu_{i}\ut_{i} )\X= \Uu \g_{\theta}(\LBD) \UT\X,
\end{equation} where $\uu_{i}$ is the i-th eigenvector, $\g_{\theta}$ is the \textit{frequency filter function} controlled by parameters $\theta$, and $l$ lowest frequency components are aggregated. The goal of $\g$ is to change the weights of eigenvalues to fit the target output. The same examples in A-1 can be rewritten in the spectral domain. For example, \textbf{GCN} can be written as
\begin{equation}\small
\Z = \An\X=(\I-\Ln)\X=\Uu  (1-\LBD) \UT\X,
\label{eq:a1_gcn2}
\end{equation}
where $\hat{\An}=\hat{\D}^{-\frac{1}{2}}\A\hat{\D}^{-\frac{1}{2}}$ is renormalization of $\An$. 
Therefore, the frequency response function is $\g(\LBD) = 1-\LBD$.
All the rewritten equations in B-0 are listed in Table \ref{tab:freq_func}.



\subsection{Polynomial of Approximation (B-2)}\label{sec:spectral_b2}
Considering higher order on eigenvalues, frequency response function is a polynomial function and thereby can approximate more complex signal compared with linear approximation.
Formally, B-2 can be written as:
\begin{equation}\small
    \Z = (\sum_{i=0}^{l} \sum_{j=0}^{k}\theta_{j}\lambda_{i}^{j} \uu_{i}\ut_{i})\X= \Uu \Pp_{\theta}(\LBD) \UT\X,
\label{eq:spectral_b2_mat}
\end{equation} where $\g(\LBD)=\Pp_{\theta}(\LBD)$ is a polynomial function of eigenvalues. 

\subsection{Rational Approximation (B-3)}\label{sec:spectral_b3}
The rational approximation is theoretically proved to be more powerful than polynomial approximation~\cite{achieser2013theory,trefethen2013approximation,petrushev2011rational,cohen2011numerical,pachon2010algorithms}, and it can handle the non-smooth signal. Rational kernel based methods are written as:  
\begin{equation}\small
    \Z =(
        \sum_{i}^{l} 
        \frac{\mathlarger\sum_{j=0}^{k}\theta_{j}\lambda_{i}^{j}}
             {\mathlarger\sum_{m=1}^{n}\phi_{m}\lambda_{i}^{m}+1}   
        \uu_{i}\ut_{i}
    )\X
    = \Uu \frac{\Pp_{\theta}(\LBD)}{\Qq_{\phi}(\LBD)} \UT\X,
    \label{eq:b3}
\end{equation}where $\g(\cdot)=\frac{\Pp_{\theta}(\cdot)}{\Qq_{\phi}(\cdot)}$ is a rational function, and $\Pp, \Qq$ are two independent polynomial functions. 



\paragraph{Remark:} Note that the examples above in B-3 all have constraints on parameters. For instance, the numerator ($\alpha$) in PPNP is equal to its bias of denominator which may render the rational function fail to obtain the optimal parameters. Therefore, one option is to learn all the separate parameters in Eqn. \ref{eq:b3}~\cite{chen2018rational}.

\section{Performance and Relationship Analysis}
As analyzed above, there is a close correspondence between the spatial and spectral methods. Also, there is a trade-off relationship between them.
In quantum mechanics, Heisenberg's uncertainty principle \cite{folland1997uncertainty} describe a limit to the accuracy for certain pairs of complementary variables or canonically conjugate variables of a particle, implying that predicting the value of a quantity with arbitrary certainty is impossible. Specifically, 
\begin{equation}\small
    \Delta_{t}^{2} \Delta_{\omega}^{2} \geq \frac{1}{4},
\end{equation}where $\Delta_{t}$ and $\Delta_{\omega}$ denote time spread and frequency spread respectively, and there is a trade-off between time and frequency concentration for a signal. Inspired by quantum uncertainty principle, spectral graph analogy is developed \cite{agaskar2013spectral}, showing a trade-off between the spatial (\textbf{A-0}) and spectral (\textbf{B-0}) domain. 

\vspace{2pt}
\noindent\textbf{Time Complexity and Expressive Power}: 
A-1/B-1 have a time complexity of $\mathcal{O}(N^2F)$ due to the matrix multiplication of $\A\X$. Accordingly, polynomial and rational method are analyzed in Table \ref{tab:analysis} where K is the order number. To compare their expressive powers, the convergence rate on challenging jump signal is employed as a benchmark~\cite{chen2018rational} (the simple signal cannot distinguish them). As shown in Table \ref{tab:analysis}, A-3/B-3 converge exponentially faster than A-1/B-1, and  A-2/B-2 converge linearly faster than A-1/B-1.
Therefore, there is a trade-off between expressive power and computational efficiency. A-1/B-1 have the best efficiency but only capture the linear relationship. A-3/B-3 consume the most considerable overhead but could tackle more challenging signals.

\begin{table}[]
\centering
\scalebox{0.89}{
\begin{tabular}{|l|l|l|l|}
\ChangeRT{2pt}
             & Linear & Polynomial & Rational \\ \ChangeRT{1pt}
Time         &  $\mathcal{O}(N^2F)$ & $\mathcal{O}(N^{K+1}F)$ & $\mathcal{O}(N^{K+1}F+N^3)$     \\ \hline
Expressivity &  $\mathcal{O}(1)$    & $\mathcal{O}(1/K)$ & $\mathcal{O}(\exp^{-\sqrt{K}})$          \\ \ChangeRT{2pt}
\end{tabular}}
\caption{Comparison on Time Complexity and Expressive Power}
\label{tab:analysis}
\end{table}

Many recent works improve GNNs by tackling over-smoothing and large scale issue. In subsections \ref{sec:sampling} and \ref{sec:over_smooth}, we will show that all the improvements are still inside our framework.

\subsection{Sampling Point of View}
\label{sec:sampling}
To handle large graph, the sampling mechanism is introduced as a spatial-based method. Popular methodologies include random walk and subgraph sampling.

\textit{DeepWalk}~\cite{perozzi2014deepwalk} first draws a group of random paths from the graph and applies a skip-gram algorithm to extract node features. Assuming the number of samples is large enough, then the transfer probability of random walk is, i.e., $\An=\D^{-1}\A$.
If the training is sufficient and samples are adequate, the node will converge to its neighbors.
Let the window size of skip-gram be $2t+1$ and the index of current node is $t+1$. Therefore, the updated representation is as follows:
\begin{equation}\small
    \Z = \frac{1}{t+1}(\I + \An + \An^{2} + ... + \An^{t})\X = \frac{1}{t+1}\Pp(\An)\X,
\end{equation}which is exactly polynomial propagation A-2.
\textit{Node2Vec}~\cite{grover2016node2vec} defines a 2nd order random walk to control the balance between Breath First Search (BFS) and Depth First Search (DFS).  
Assuming the random walk is sufficiently sampled, Node2Vec can be rewritten in matrix form: 
\begin{equation}\footnotesize
    \Z = (\frac{1}{p}\cdot\overbrace{\I}^\text{source}+\overbrace{\An}^\text{BFS}+\frac{1}{q}\overbrace{(\An^{2}-\An)}^\text{DFS})\X,
\label{eq:node2vec_mat}
\end{equation} 
where transition probabilities $\An=\Drw\A$ is random walk normalized adjacency matrix, and Node2Vec also  belongs to A-2.
\textit{FastGCN}~\cite{chen2018fastgcn} and \textit{GraphSAGE}~\cite{hamilton2017inductive} sample subgraph randomly, which is equivalent to employing important sampling for each batch. The transition probability follows random walk normalization ($\An=\D^{-1}\A$), making it belong to A-1.

\vspace{2pt}
\noindent\textbf{Remark}: No sampling method belongs to A-3 or B-3. This is reasonable since A-3/B-3 has dramatically higher computational complexity.

\subsection{Over-smoothing Point of View} 
\label{sec:over_smooth}
Most GNNs perform poorly when stacking many layers, which is called the over-smoothing issue.
Many related works aiming to solve the over-smoothing issue can be reduced to one category of the proposed framework.

\cite{zhu2020beyond} proposed a method that combines direct neighbors and higher-order, which equivalent to polynomial propagation (A-2) or approximation (B-2). 
Deep GCN~\cite{li2019deepgcns} developed a model with a residue module, dense connection, and dilated aggregation, which learns the weights of all different orders of neighbors. This is equivalent to polynomial propagation (A-2) or approximation (B-2). JKNet~\cite{xu2018representation} also follows the same residue methodology as Deep GCN.
PairNorm~\cite{zhao2019pairnorm} presents a two-step method that includes centering and re-scaling, which can neutralize the aggregation from graph convolution. PairNorm is similar to rational propagation (A-3) or approximation (B-3) since re-scaling is similar to teleport or restart operation.
\cite{bo2021beyond} design an adaptive method to dynamically adjust the weights between low-frequency and high-frequency components, resulting in two peaks in the spectral domain. This could also be modeled by rational propagation (A-3) or approximation (B-3) with its accuracy in jump signals.
DropEdge~\cite{rong2019dropedge} randomly drops a certain number of edges to avoid over-smoothing, which can be categorized as rational propagation (A-3) or approximation (B-3) since dropping edge provides a probability of keeping the original values of nodes. 

\vspace{2pt}
\noindent\textbf{Remark}: No stat-of-the-art method belongs to A-1 or B-1, which implies that A-1 and B-1 are fragile to the over-smoothing issue.

\section{Conclusion}
This paper proposes a unified framework that summarizes the state-of-the-art GNNs, providing a new perspective for understanding GNNs with different mechanisms. By analytically categorizing current GNNs into the spatial and spectral domains and further dividing them into subcategories, our analysis reveals that the subcategories are strongly connected by generalization, specialization, and equivalence relation. Even for currently trending topics such as over-smoothing studies, our proposed method can generalize them as one of those subcategories.

\newpage
\bibliographystyle{named}
\bibliography{ijcai21}

\end{document}